\documentclass{article}
\usepackage{graphicx}
\usepackage{framed,multirow}
\usepackage[letterpaper, portrait, margin=1in]{geometry}

\usepackage[shortlabels]{enumitem}

\providecommand{\keywords}[1]
{
  \small	
  \textbf{\textit{Keywords---}} #1
}

\title{Detecting Suspicious Behavior: How to Deal with Visual Similarity through Neural Networks}
\author{
  Mart\'{i}nez-Mascorro, Guillermo A.\\
  \texttt{a00824126@itesm.mx}
  \and
  Ortiz-Bayliss, Jos\'{e} C.\\
  \texttt{jcobayliss@tec.mx}
  \and
  Terashima-Mar\'{i}n, Hugo\\
  \texttt{terashima@tec.mx}
}
\date{}

\begin{document}

\maketitle

\begin{abstract}
    Suspicious behavior is likely to threaten security, assets, life, or freedom. This behavior has no particular pattern, which complicates the tasks to detect it and define it. Even for human observers, it is complex to spot suspicious behavior in surveillance videos. Some proposals to tackle abnormal and suspicious behavior-related problems are available in the literature. However, they usually suffer from high false-positive rates due to different classes with high visual similarity. The Pre-Crime Behavior method removes information related to a crime commission to focus on suspicious behavior before the crime happens. The resulting samples from different types of crime have a high-visual similarity with normal-behavior samples. To address this problem, we implemented 3D Convolutional Neural Networks and trained them under different approaches. Also, we tested different values in the number-of-filter parameter to optimize computational resources. Finally, the comparison between the performance using different training approaches shows the best option to improve the suspicious behavior detection on surveillance videos.
\end{abstract}


\section{Introduction}
\label{sec:Introduction}

    Suspicious behavior detection has become an important and exciting topic due to its potential for several applications~\cite{SuspiciousFinancialTransactionsTang2012, StorageBasedPennington2010, AnomalySurveyVarun2009}. Recent years have witnessed a growing interest in developing automatic detection methods to improve security and surveillance systems~\cite{ContextSpaceModelWiliem2012}. However, defining what ``suspicious behavior" is, from a human perspective, is complicated since this behavior lacks a particular pattern, nor a fixed set of actions. Consequently, it is challenging to recognize such behavior accurately, even for human observers. People working in surveillance tasks suggest that suspicious behavior detection requires personal understanding and subjective interpretation~\cite{ContextSpaceModelWiliem2012, ComportamientoNoVerbalRafael2016}. Human observers rely on their instincts, but it takes several years of practice and experience to develop this ``gut feeling"~\cite{CrimeCCTVAustraliaWells2006}.
    
    Many behavioral specialists, police departments, and foundations agree on describing several suspicious behaviors for better comprehension~\cite{ADLFoundation, BerwynPoliceDepartment, HowStuffWorks}. According to the location where the suspicious behavior takes place, it may have different interpretations, such as crime commission~\cite{MetropolitanPoliceDepartment}, terrorism~\cite{HomeLandSecurity}, campus security~\cite{UniversityOfMichiganDPSS}, among others. As stated by the Hilliard Police Department~\cite{HilliardPoliceDepartment}, people are not suspicious, but their behavior. In other words, it is the behavior that matters, and not particular skin color, typical clothes, or facial expressions; it is how a person behaves plus several details. Although it is hard to explain, most of the security and prevention associations try to involve the community to spot suspicious behaviors. The earlier those conducts are found, the less damage they can do.
        
    Suspicious behavior detection systems are developed for particular situations and specific behaviors in mind. Usually, these systems are related to crime commission or prevention. For example, Tang and He~\cite{SuspiciousFinancialTransactionsTang2012} looked for suspicious financial transactions to prevent money laundering, Pennington et al.~\cite{StorageBasedPennington2010} examined storage-data access patterns to prevent intrusion and data theft, and Penmetsa et al.~\cite{UAVSurveillancePenmetsa2014} proposed aerial visual surveillance to detect violent actions, such as shooting, hitting, or choking. 
    
    Although many works have focused on crime scene detection~\cite{UCFCrimes2018} and abnormal behavior detection, such as shooting, robbery, or car accidents, only a few works focus on the behavior before the crime commission. More importantly, those few works provide no specific details about the type of offense. In this regard, it is important to mention the Pre-Crime Behavior~(PCB) method~\cite{PCBand3DCNN}, which has been applied to suspicious behavior detection in shoplifting cases. After using the PCB method and eliminating most of the crime-details information, normal and suspicious samples look very similar. Then, classifiers tend to produce high false-positive rates the visual similarity between different classes~\cite{SuspBehaviorHDMu2016,WeaklySupervisedHu2020}. A question arises from this situation: if suspicious behavior is generalized within the crime prevention context, should suspicious-behavior samples from different crimes be trained as one class or separately?
           
    In this work, we present a comparison between three models that combine training-and-classification approaches for suspicious behavior detection. The first model relies on binary training and classification, which can only sort input samples as normal or suspicious. The second model relies on multi-class training and ends with a binary classification or a multi-class classification that discriminates among five classes. To test such models, we used 1,278 samples videos from the complete UCF-Crime dataset~\cite{UCFCrimes2018}, processed with PCB, which includes normal-behavior and four types of crimes: shoplifting, stealing, arson, and abuse. Also, for each model, we tried six different number-of-filters configurations. The main contribution derived from this work is that grouping suspicious-behavior samples while training, regardless of the crime, improves the classification accuracy. In an overall comparison, results show a 93.4\% accuracy for the binary model (2.5\% better than the multi-class model).
        
    The remainder of this document is organized as follows. The most relevant works related to this investigation are briefly introduced in Section~\ref{sec:RelatedWork}. In Section~\ref{sec:Experiments}, we present the methods developed as well as the experimental design. Section~\ref{sec:Results} presents and discusses the results obtained. Finally, the conclusions and future work are presented in Section~\ref{sec:Conclusion}.
        
    \begin{figure*}
    \centering
        \includegraphics[width = \textwidth]{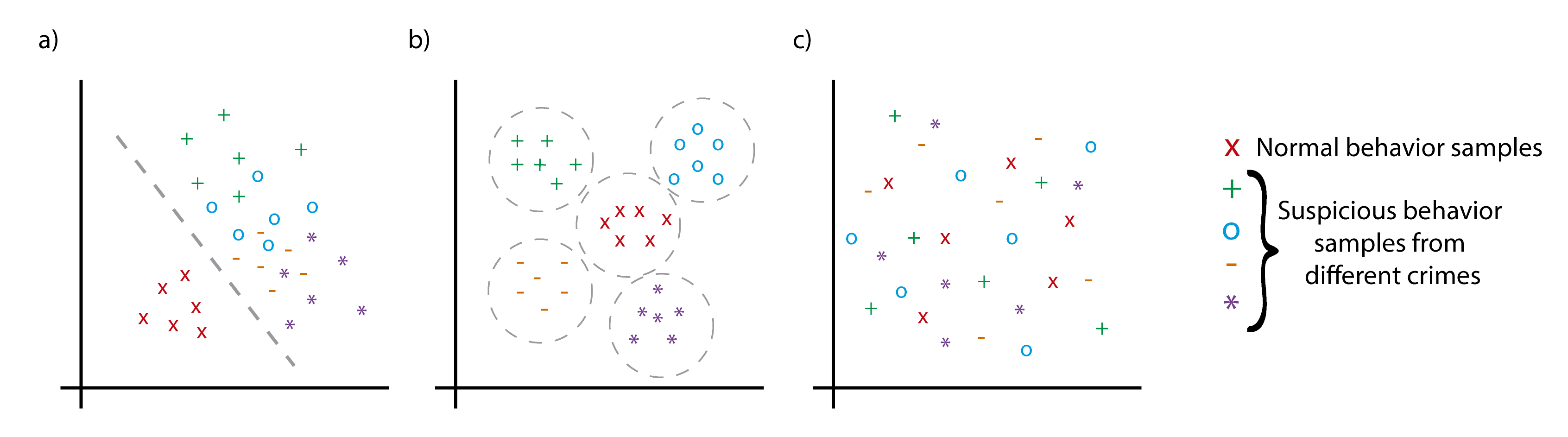}
        \caption{Possible samples-distribution scenarios.}
        \label{fig:SamplesDistributionAssumption}
    \end{figure*}

\section{Related work}
\label{sec:RelatedWork}

    Suspicious behavior is usually confused with abnormal behavior, but they have different notions. Suspicious behavior regularly refers to unusual interactions between people or people and the objects around them~\cite{SuspBehaviorHDMu2016}. Abnormal or anomalous behavior usually refers to everything outside the usual or expected behavior~\cite{AnomalySurveyVarun2009, SuspiciousPathsVaswani2003, AnomalousDetectionJiang2011, AnomalyDetectionSabokrou2016}. Works on abnormal behavior detection focus on modeling normal conduct, and everything the model cannot classify as `usual' is considered abnormal.

    Abnormal behavior detection systems are the predecessors and foundation for suspicious behavior detection systems. Some examples of abnormal behavior detection applications include abnormal motion~\cite{AbnormalMotionHuan2011}, abnormal trajectories~\cite{PedestrianAbnormalWang2018}, and abnormal crowd behavior~\cite{CrowdBehaviorTripathi2019}. However, some behaviors could be unusual or abnormal without being suspicious, such as a small group reunion in the street, or waiting for someone. Taking preventive measures against them could be considered discriminatory or even illegal~\cite{SuspiciousPositionalRowe2005}. Some studies about non-verbal behavior indicate that it is highly likely for anyone to fail on human behavior understanding without considering the context where they observed~\cite{BodyMovementBull1983, ComportamientoNoVerbalRafael2016}. In other words, an observer could consider a behavior suspicious in a particular context, but reasonable in a different situation.
    
    Conversely, most of the available approaches for detecting suspicious behavior try to prevent specific scenarios, mostly related to crimes. For example, Rowe~\cite{SuspiciousPositionalRowe2005} presented a suspicious-behavior detection system based on wireless sensors and changes in positions, velocities, and accelerations. The system considered walking paths in a room a, looking or suspicious patterns and unusual spots to stop. Jun and Lei~\cite{SuspiciousFinancialTransactionsTang2012} combined genetic algorithms and backpropagation neural networks to detect suspicious behavior on financial transactions. They used the genetic algorithm to find the better initial weights for the network, and a dataset from a commercial bank. Goodall et al.~\cite{SituGoodall2019} introduced a tool to fight against cybercrime. This tool supports operators to discover unusual behaviors in streaming network data. The system parses several event streams and scores each of them, and in a second stage, it provides visual support for the analyst to explore and understand the context.
    
    Also, many works aim to support CCTV surveillance tasks and prevent crimes such as robbery, shoplift, riots, among others. For example, Mu et al.~\cite{SuspBehaviorHDMu2016} proposed a recognition algorithm for high definition videos based on motion vectors. Their algorithm processes video samples of $1920 \times 1080$ pixels and uses a macroblock of $4 \times 4$. They build a video dataset, which includes behaviors such as wandering, trailing, chasing, falling, and normal activity. Xia et al.~\cite{CaseBasedXia2015} presented a case-based reasoning approach, using a saliency-based visual attention model combined with time-attribute features. They proposed to decompose the general behavior into sub-behaviors and estimate every body position. Using Hidden Markov models, they recognize each action and take initial and end times. Mart\'{i}nez et al.~\cite{PCBand3DCNN} presented a new approach to process criminal video samples. They proposed the Pre-Crime Behavior method~(PCB) to separate a crime-commission sample in three segments and extract a known-offender behavior before even acting suspiciously. Combining PCB and 3D Convolutional Neural Networks~(3DCNN), the model detected suspicious behavior with 75\% accuracy in shoplifting samples. 
    
    Suspicious and normal samples have a high visual similarity after applying the PCB method. Fig~\ref{fig:SamplesDistributionAssumption} shows three possible scenarios:
    
    \begin{enumerate}[a)]
        \item A hyperplane can easily split suspicious and normal samples.
        \item A single hyperplane cannot separate the classes, so it is better to consider a multi-class approach.
        \item The samples are very similar and cannot be separated.
    \end{enumerate}
    
    Convolutional Neural Networks~(CNN) have a high performance in computer vision and pattern recognition. Many approaches implemented them to tackle problems such as object detection~\cite{ObjectDetectionLee2018, IODEum2017}, identifying actions in images~\cite{GAPActionRecognitionWang2018}, and text recognition~\cite{HandwrittenTextNaur2018}. Ji et al.~\cite{First3DCNNJi2013} proposed a three-dimensional convolution on a CNN (3DCNN) architecture to analyze video data. This new approach allowed them o extract spatial and temporal features and opened new analysis areas such as anomaly detection~\cite{DeepCascadeSabokrou2017}, gesture recognition~\cite{ZhangGestures3DCNN2017} and Magnetic Resonance Imaging (IMR) analysis~\cite{MRIAgeStimationUeda2019,MRIHyperactivityZou2017, MRIAlzheimerKhagi2018}.
    
    Considering these scenarios and types of samples, we propose to explore the visual similarity between behavior samples, after using the PCB method. Section~\ref{sec:Experiments} presents the details about the dataset, the 3DCNN model, and the approaches to train it.
    
\section{Experiments}
\label{sec:Experiments}

    In this investigation, we compared two training approaches by conducting two different experiments. Both experiments dealt with 3D Convolutional Neural Networks~ (3DCNN) trained for suspicious behavior detection. Also, they considered six configurations for the number of filters in the convolutional layers to evaluate the necessary input decomposition. 

    The first experiment dealt with a binary training approach. We assumed that suspicious-behavior samples from different crime videos are similar and easily separable from normal-behavior samples (Fig.~\ref{fig:SamplesDistributionAssumption}, a). The experiment includes thirty 3DCNN trained for each filter-configuration and using the binary training approach.
    
    The second experiment tested a multi-class training approach. In this case, we assumed that a single hyperplane could not separate the suspicious-behavior samples from themselves, either the normal-behavior samples (Fig.~\ref{fig:SamplesDistributionAssumption}, b). We tested multi-class and binary classification from the multi-class trained 3DCNN. Combinations between training and classification approaches are explained in subsection~\ref{subsec:TrainningApproaches}. Also, this experiment combined the training approach with six number-of-filter configurations and performed thirty runs for each combination.
    
    We used crime video samples for the experiments and processed them with the PCB method. Then, we trained 360 different 3DCNN using two approaches: binary and multi-class. We also modified the number of filters' parameter by each convolutional layer, to look for the necessary decomposition for this task.

    \subsection{Dataset}
    \label{subsec:Dataset}

        The generated dataset includes 1,278 video samples based on the UCF-Crime Dataset~\cite{UCFCrimes2018}. The samples include shoplifting, stealing, abuse, arson, and normal behavior classes. The dataset contains a fraction of the videos from each class. The selection process included videos where at least one person's behavior is visible before the crime is committed. Also, we processed all the videos with the PCB method and resized the video resolution to $80 \times60$ pixels since this resolution showed a good performance in previous comparisons~\cite{PCBand3DCNN}.

    \subsection{The Pre-Crime Behavior Method}
    \label{subsec:PCB}
    
\begin{figure}[htb]
        \centering
            \includegraphics[width = .8\linewidth]{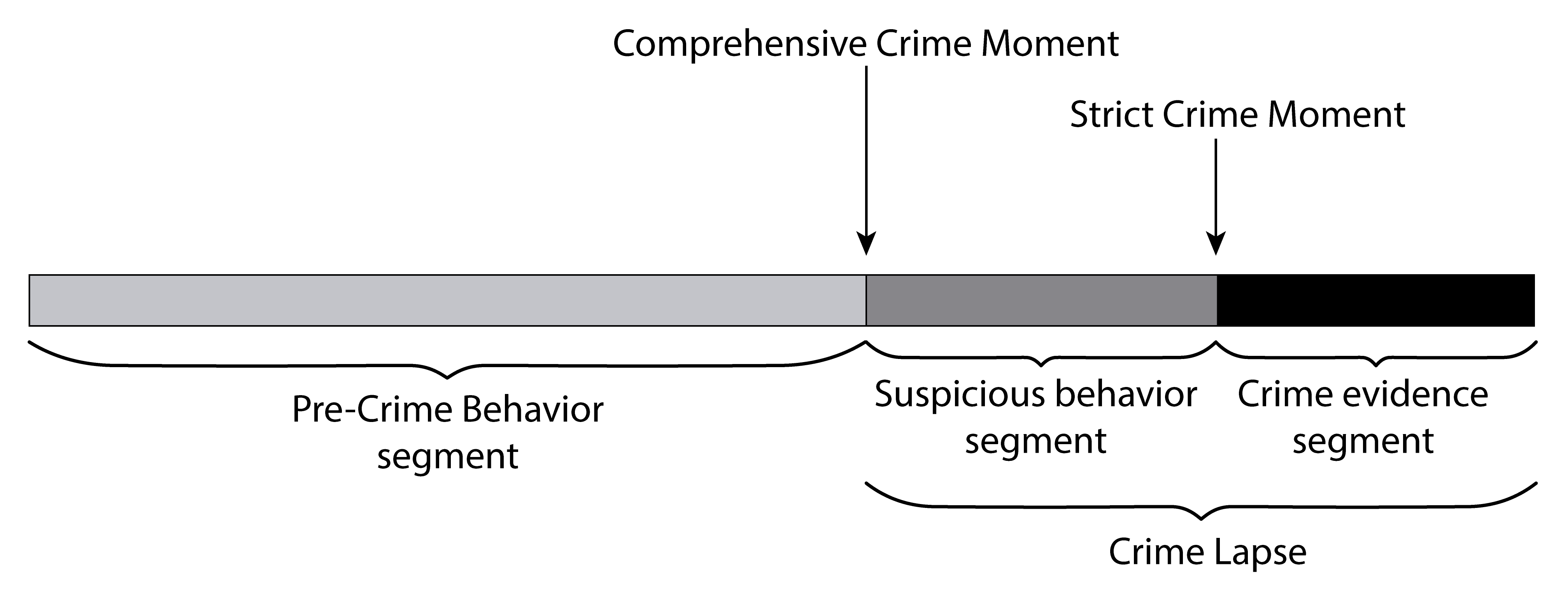}
            \caption{PCB process representation.}
            \label{fig:PCBprocess}
        \end{figure}

        The PCB method allows processing video samples where an observer can see an offender's behavior before committing a crime. The method splits the video into three parts (Fig.~\ref{fig:PCBprocess}): the crime evidence segment, the suspicious behavior segment, and the pre-crime behavior segment. To process the videos, the observer must watch the complete sample and detect specific moments. The crime evidence segment starts where the suspect in the video unquestionably commits an offense, denominated Strict Crime Moment~(SCM). The suspicious behavior segment begins where the observer doubts about the person in the video (Comprehensive Crime Moment, CCM). At this point, the suspect is behaving unusually. The third segment is what occurs from the first appearance of the suspect, and just before he started to act suspicious.
        
        PCB segments are like looking for regular clients in a store, to human sight. They only show people walking around the area and looking for products. These segments lack enough information to raise suspicion about a specific person. Therefore, they present a high visual similarity with normal-behavior samples.
        
        We looked for a significant difference between suspicious-behavior and normal-behavior samples using the PCB segments to train a 3DCNN model. If the difference is substantial, the combination may work for suspicious behavior detection, even before a human observer can recognize it.
                
    \subsection{3D Convolutional Neural Network}
    \label{subsec:3DCNN}
            
        Convolutional Networks have shown a remarkable performance in object and action detection. Many works have proposed approaches to evolve these networks to analyze a sequence of actions and look for specific behaviors. 3D Convolutional Neural Networks~(3DCNN) are capable of processing spatial-temporal information, such as video samples, and extract significant features.
        
\begin{figure}[b!]
        \centering
            \includegraphics[width = 0.8\linewidth]{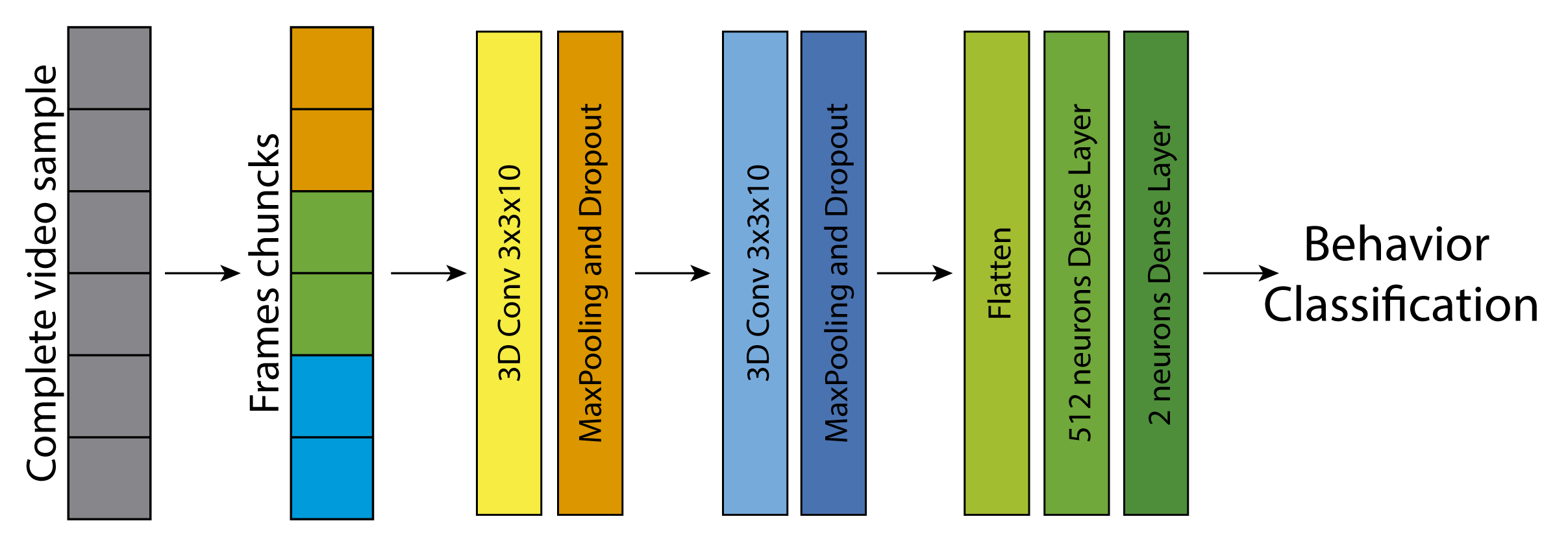}
            \caption{3DCNN architecture}
            \label{fig:3DCNN}
        \end{figure}

        Although several works developed particular architectures to solve specific problems~\cite{Ref1ConvArchitecture2018, Ref3ConvArchitectures2019, DeepCascadeSabokrou2017}, we implemented a basic 3DCNN to test visual similarity between normal-behavior samples and suspicious-behavior samples processed with the PCB method. The architecture of the model consists of two Conv3D layers, two max-pooling layers, and two fully connected layers (Fig~\ref{fig:3DCNN}).
            
    \subsection{Training and Classification Approaches}
    \label{subsec:TrainningApproaches}
        
\begin{figure}[htb]
            \centering
            \includegraphics[width = 0.7\linewidth]{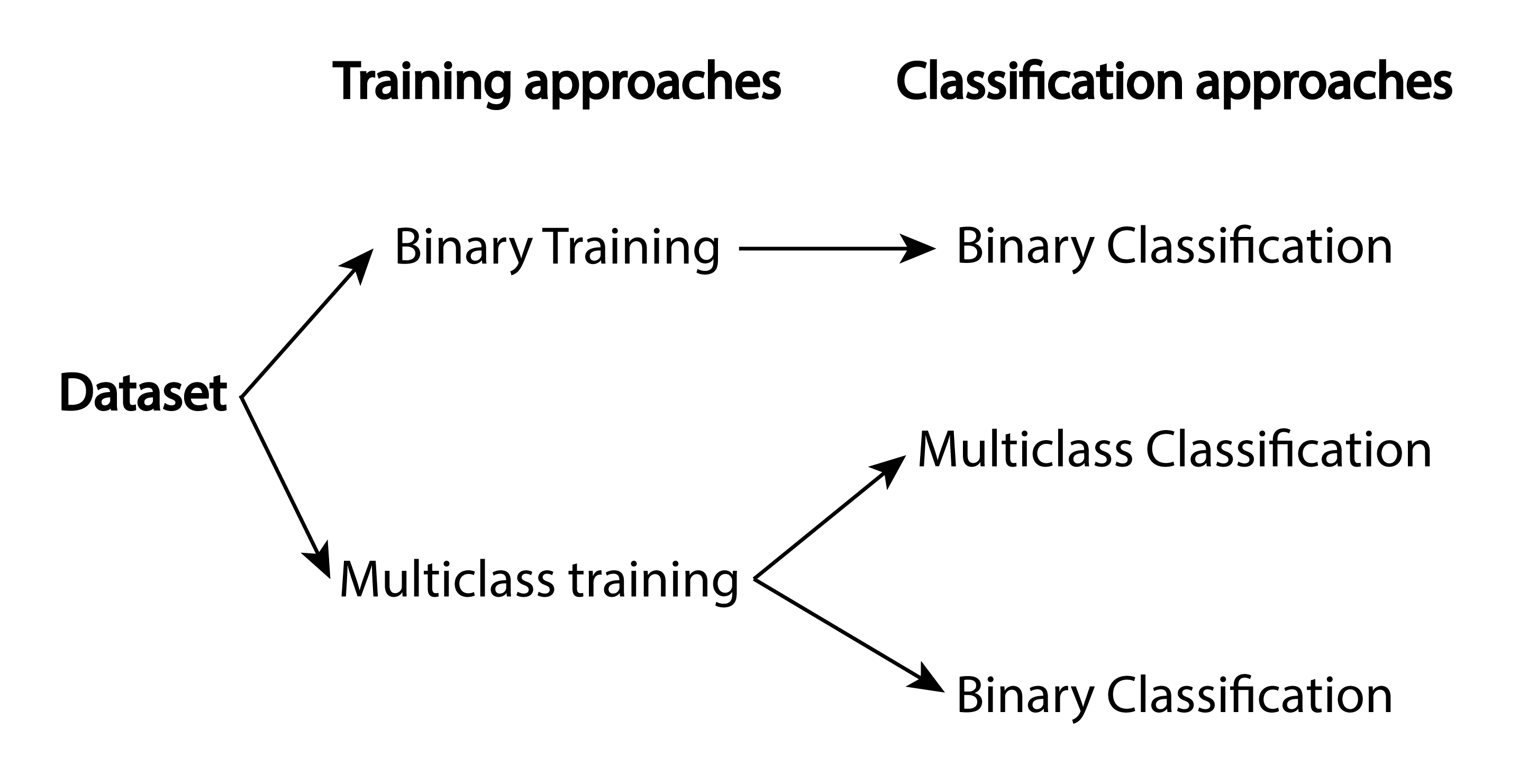}
            \caption{Training and classification approaches}
            \label{fig:TrainingApproaches}
        \end{figure}

        We choose binary and multiclass approaches for training and classification stages (Fig.~\ref{fig:TrainingApproaches}). These approaches considered the sample's distribution from Fig.~\ref{fig:SamplesDistributionAssumption} a) and b). Binary training works with binary classification, but from multi-class training, we tested binary classification and multi-class classification. Binary classification from multi-class training considers as a positive any crime sample classified in any crime class, e.g., if the model classifies a shoplifting sample in the arson class, it counts as a suspicious-behavior detection. 
        
        The experiment considers six couples of number-of-filters for each convolutional layer: 16-16, 32-32, 32-64, 64-64, 64-128, 128-32. As the higher number of filters, the more details the network extracts, but also, the more computational resources it consumes. We looked for the best couple of values to achieve a useful classification and optimizing resources.
        
        We run each configuration 30 times to validate the results. Due to the unbalance in the dataset, we consider the balanced accuracy~(bACC). It normalizes true positive~(TP) and true negative~(TN) predictions by the number of positive and negative samples, respectively. The balanced accuracy is defined in (\ref{eq:bACC}).
        
        \begin{equation}
            bACC = \frac{TPR + TNR}{2}
            \label{eq:bACC}
        \end{equation}
                
        \begin{equation}
            TPR = \frac{TP}{TP+FN}
            \label{eq:TPR}
        \end{equation}
                
        \begin{equation}
            TNR = \frac{TN}{TN+FP}
            \label{eq:TNR}
        \end{equation}
                
        Where TPR is the True Positive Rate~(\ref{eq:TPR}), TNR is the True Negative Rate~(\ref{eq:TNR}), TP is True Positive, TN is True Negative, FP is False Positive, and FN is False Negative predictions from the confusion matrix.

\section{Results}
\label{sec:Results}
        
\begin{figure}[htb]
    \centering
        \includegraphics[width = .75\linewidth]{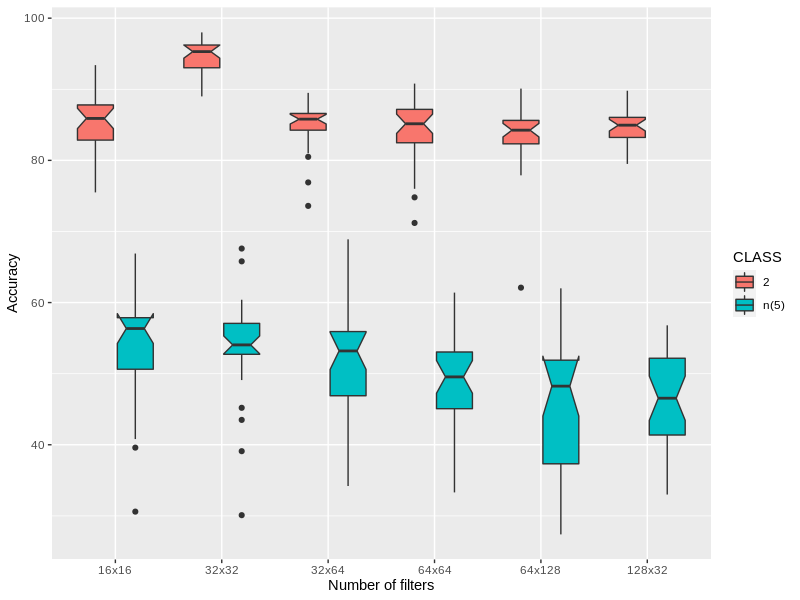}
        \caption{Balanced accuracy comparison between binary training-binary classification approach and multi-class training-multi-class classification approach, using six filter combinations.}
        \label{fig:bAccUnfair}
    \end{figure}

    After getting the balanced accuracy of all models, we first compared the binary classification against the multi-class classification. The first comparison was between the binary classification against the multi-class classification. Although we looked for suspicious-behavior detection, we tested the capability of the multi-class training to detect each crime by separate. We performed a t-test on the means for each pair of configurations (Fig.~\ref{fig:bAccUnfair}). The test had two hypotheses: H0, where both means are equal, and H1, where means are not. Following the standard, we assumed a significance value of 5\% (alpha = 0.05). The results showed overwhelming statistical evidence in favor of binary training-binary classification. Table~\ref{tab:Pvalues} shows the p-values of each test, and they point out that the means are not equal in all cases.
    
    \begin{table}[b]
    \centering
        \caption{P-values from t-test. Comparison between 2-classes against 5-classes classification.}
        \label{tab:Pvalues}
            \begin{tabular}{cc}
                \hline
                \multicolumn{1}{l}{\textbf{Filter pair}} & \multicolumn{1}{l}{\textbf{p-value}} \\ \hline
                16 - 16 & $3.68e^{-19}$ \\
                32 - 32 & $1.39e^{-23}$ \\
                32 - 64 & $1.38e^{-20}$ \\
                64 - 64 & $3.51e^{-21}$ \\
                64 - 128 & $1.65e^{-17}$ \\
                128 - 32 & $4.88e^{-23}$ \\ \hline
            \end{tabular}
    \end{table}
    
    In a second comparison between binary classification models ---one from binary training and other from multi-class training---, the multi-class training-binary classification approach did not exceed the binary training-binary classification approach even though the p-values increase. In all cases, as the p-values are lower than alpha, H0 is rejected. 

    Table~\ref{tab:BestBACC} presents the balanced accuracy of the tested approaches. The first approach --- binary training and binary classification --- shows a better performance from the other two. The difference against the second approach is overwhelming, while the third approach is 2.5\%. To be fair, the binary training approach has a better performance but cannot distinguish between the different types of crimes. 
    
    It is important to remember that this work aims to generalize suspicious behavior from different types of crimes. If the goal is to detect or prevent a specific kind of crime, it will be necessary to improve the multi-class detection approach. Also, the three methods present their better results using 16 filters on both convolutional layers. This insight may support the development of a real-time detection app and in more agile neural network trainings.
    
    \begin{figure}[htb]
    \centering
        \includegraphics[width = .75\linewidth]{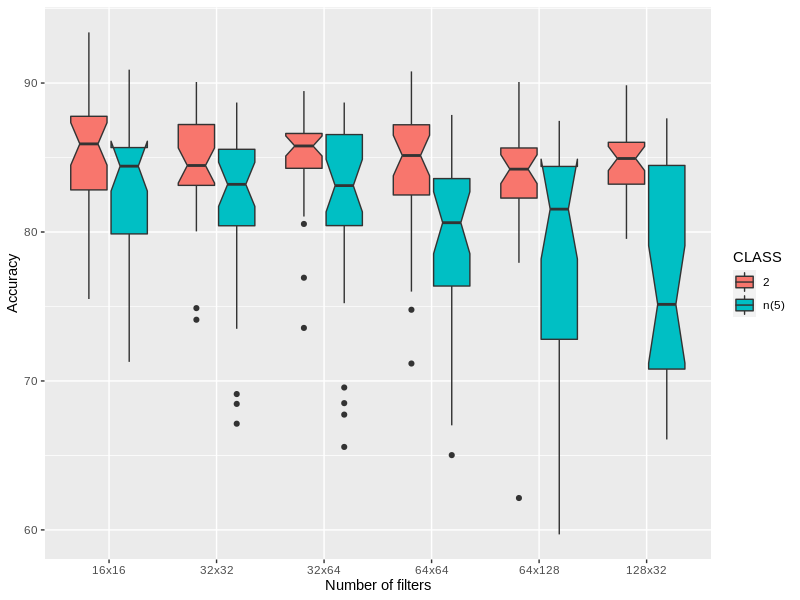}
        \caption{Balanced accuracy comparison between binary training-binary classification approach and multi-class training-binary classification approach, using six filter combinations.}
        \label{fig:Results}
    \end{figure}
        
    \begin{table}[b!]
    \centering
    \caption{P-values from t-test. Comparison between binary classification.}
    \label{tab:PvaluesTest2}
        \begin{tabular}{cc}
            \hline
            \multicolumn{1}{l}{\textbf{Filter pair}} & \multicolumn{1}{l}{\textbf{p-value}} \\ \hline
            16 - 16 & $3.06e^{-2}$ \\
            32 - 32 & $3.20e^{-2}$ \\
            32 - 64 & $2.87e^{-2}$ \\
            64 - 64 & $6.79e^{-4}$ \\
            64 - 128 & $8.99e^{-3}$ \\
            128 - 32 & $1.54e^{-5}$ \\ \hline
        \end{tabular}
    \end{table}
        
    \begin{table*}[t!]
        \centering
        \caption{Best balanced accuracy (bACC) by each training approach and number-of-filter values.}
        \label{tab:BestBACC}
        \resizebox{\textwidth}{!}{%
        \begin{tabular}{cccc}
        \hline
        \textbf{\begin{tabular}[c]{@{}c@{}}Number-of-filter\\ values\end{tabular}} & \textbf{\begin{tabular}[c]{@{}c@{}}Binary training\\ Binary classification (\%)\end{tabular}} & \textbf{\begin{tabular}[c]{@{}c@{}}Multiclass training\\ Multi-class classification (\%)\end{tabular}} & \textbf{\begin{tabular}[c]{@{}c@{}}Multiclass training\\ Binary classification (\%)\end{tabular}} \\ \hline
        \textbf{16 - 16} & 93.4 & 68.9 & 90.9 \\
        \textbf{32 - 32} & 90.1 & 67.6 & 88.7 \\
        \textbf{32 - 64} & 89.5 & 68.9 & 88.7 \\
        \textbf{64 - 64} & 90.8 & 61.4 & 87.9 \\
        \textbf{64 - 128} & 90.1 & 62.0 & 87.5 \\
        \textbf{128 - 32} & 89.8 & 56.8 & 87.6 \\ \hline
        \end{tabular}
        }
    \end{table*}
    
\section{Conclusion}
\label{sec:Conclusion}

    This work presented a comparison between two training approaches for suspicious behavior detection from different types of crime processed with the Pre-Crime Behavior method. As this method removes significant information to detect crime in a video, the resulting samples have a high-visual similarity with normal-behavior samples. We looked for the best approach to avoid this similarity and classify the suspicious and normal behaviors correctly.
    
    The multiclass training approach leads to multiclass and binary classification. The neural network trained each class by separate for the first classification approach. For the binary approach, we combine the suspicious behavior samples from different types of crimes in a single class.
    
    The first comparison was between binary training-binary classification and multi-class training-multi-class classification. The results showed an immense difference, binary classification from binary training achieves better performance (24.5\% comparing the best models) than the second approach.
        
    The second multi-class approach improves by reducing the classification from five to two classes. In this scenario, we consider as positive any suspicious-behavior sample classified in one of the four crime classes. After thirty runs of the model with both approaches and getting the balanced accuracy, we perform a t-test on each pair of configurations. The results showed significance values to consider that the means are not equal, and demonstrate that the first approach achieves a better performance again.
    
    As mentioned, the binary training approach cannot identify a particular type of crime. This work aims to determine if suspicious behavior samples should be trained as a group, regardless of the nature of the offense. If the goal were to detect or prevent a particular type of crime, it would be necessary to improve the multi-class training approach or select a different one from the literature.
        
    As future work, we look for a generalization test by training with less criminal classes and testing with complete unknown ones. Also, it will be interesting to test this approach to develop real-time detection for surveillance support. We consider it essential to collaborate with behavioral analysts from different areas, such as police officers, casino staff, behavioral experts, among others, to improve the project performance.

\bibliographystyle{unsrt}
\bibliography{references}

\begin{thebibliography}{10}

\bibitem{SuspiciousFinancialTransactionsTang2012}
J.~{Tang} and L.~{He}.
\newblock Genetic optimization of bp neural network in the application of
  suspicious financial transactions pattern recognition.
\newblock In {\em 2012 International Conference on Management of e-Commerce and
  e-Government}, pages 280--284, 2012.

\bibitem{StorageBasedPennington2010}
Adam~G. Pennington, John~Linwood Griffin, John~S. Bucy, John~D. Strunk, and
  Gregory~R. Ganger.
\newblock Storage-based intrusion detection.
\newblock {\em ACM Trans. Inf. Syst. Secur.}, 13(4), December 2010.

\bibitem{AnomalySurveyVarun2009}
Varun Chandola, Arindam Banerjee, and Vipin Kumar.
\newblock Anomaly detection: A survey.
\newblock {\em ACM Comput. Surv.}, 41(3), July 2009.

\bibitem{ContextSpaceModelWiliem2012}
Arnold Wiliem, Vamsi Madasu, Wageeh Boles, and Prasad Yarlagadda.
\newblock A suspicious behaviour detection using a context space model for
  smart surveillance systems.
\newblock {\em Computer Vision and Image Understanding}, 116(2):194 -- 209,
  2012.

\bibitem{ComportamientoNoVerbalRafael2016}
Rafael Manuel~L{\'o}pez P{\'e}rez, Fernando~Gordillo Le{\'o}n, and Marta~Grau
  Olivares.
\newblock {\em Comportamiento no verbal: m{\'a}s all{\'a} de la
  comunicaci{\'o}n y el lenguaje}.
\newblock P{\'\i}ramide, 2016.

\bibitem{CrimeCCTVAustraliaWells2006}
Helene Wells, Troy~John Allard, and Paul Wilson.
\newblock Crime and cctv in australia: Understanding the relationship.
\newblock In {\em Political Science}, 2006.

\bibitem{ADLFoundation}
{Anti-Defamation League (ADL)}.
\newblock Recognizing and dealing with suspicious people.
\newblock accessed May 4th 2020.

\bibitem{BerwynPoliceDepartment}
{Berwyn Police Department}.
\newblock What is suspicious activity?, 2012.
\newblock accessed May 4th 2020.

\bibitem{HowStuffWorks}
D.~Ross.
\newblock Defining `suspicious behavior' without bias is harder than you think,
  2018.
\newblock accessed May 4th 2020.

\bibitem{MetropolitanPoliceDepartment}
{Metropolitan Police Department}.
\newblock Capital watch: What is suspicious behavior?
\newblock accessed May 4th 2020.

\bibitem{HomeLandSecurity}
{Homeland Security}.
\newblock What is suspicious activity?
\newblock accessed May 4th 2020.

\bibitem{UniversityOfMichiganDPSS}
{University of Michigan DPSS}.
\newblock Reporte a crime or concern suspicious behavior.
\newblock accessed May 4th 2020.

\bibitem{HilliardPoliceDepartment}
{Hilliard Police Department}.
\newblock What is a suspicious person?
\newblock accessed May 4th 2020.

\bibitem{UAVSurveillancePenmetsa2014}
Surya Penmetsa, Fatima Minhuj, Amarjot Singh, and SN~Omkar.
\newblock Autonomous uav for suspicious action detection using pictorial human
  pose estimation and classiﬁcation.
\newblock {\em ELCVIA Electronic Letters on Computer Vision and Image
  Analysis}, 13(1):18--32, 2014.

\bibitem{UCFCrimes2018}
Waqas Sultani, Chen Chen, and Mubarak Shah.
\newblock Real-world anomaly detection in surveillance videos.
\newblock {\em 2018 IEEE/CVF Conference on Computer Vision and Pattern
  Recognition}, pages 6479--6488, 06 2018.

\bibitem{PCBand3DCNN}
Guillermo~A. Martínez-Mascorro, José~R. Abreu-Pederzini, José~C.
  Ortiz-Bayliss, and Hugo Terashima-Marín.
\newblock Suspicious behavior detection on shoplifting cases for crime
  prevention by using 3d convolutional neural networks.
\newblock {\em arXiv:2005.02142v1 [cs.CV]}, 2020.

\bibitem{SuspBehaviorHDMu2016}
Chundi Mu, Jianbin Xie, Wei Yan, Tong Liu, and Peiqin Li.
\newblock A fast recognition algorithm for suspicious behavior in high
  definition videos.
\newblock {\em Multimedia Syst.}, 22(3):275–285, June 2016.

\bibitem{WeaklySupervisedHu2020}
Xing Hu, Jian Dai, Yingping Huang, Haima Yang, Liang Zhang, Wenming Chen, Genke
  Yang, and Dawei Zhang.
\newblock A weakly supervised framework for abnormal behavior detection and
  localization in crowded scenes.
\newblock {\em Neurocomputing}, 383:270 -- 281, 2020.

\bibitem{SuspiciousPathsVaswani2003}
N.~{Vaswani}, A.~{Roy Chowdhury}, and R.~{Chellappa}.
\newblock Activity recognition using the dynamics of the configuration of
  interacting objects.
\newblock In {\em 2003 IEEE Computer Society Conference on Computer Vision and
  Pattern Recognition, 2003. Proceedings.}, volume~2, pages II--633, 2003.

\bibitem{AnomalousDetectionJiang2011}
Fan Jiang, Junsong Yuan, Sotirios~A. Tsaftaris, and Aggelos~K. Katsaggelos.
\newblock Anomalous video event detection using spatiotemporal context.
\newblock {\em Comput. Vis. Image Underst.}, 115(3):323–333, March 2011.

\bibitem{AnomalyDetectionSabokrou2016}
M.~{Sabokrou}, M.~{Fathy}, and M.~{Hoseini}.
\newblock Video anomaly detection and localisation based on the sparsity and
  reconstruction error of auto-encoder.
\newblock {\em Electronics Letters}, 52(13):1122--1124, 2016.

\bibitem{AbnormalMotionHuan2011}
Ruo~Hong Huan, Xiao~Mei Tang, Zhe~Hu Wang, and Qing~Zhang Chen.
\newblock Abnormal motion detection for intelligent video surveillance.
\newblock In {\em Information Technology for Manufacturing Systems II},
  volume~58 of {\em Applied Mechanics and Materials}, pages 2290--2295. Trans
  Tech Publications, 2011.

\bibitem{PedestrianAbnormalWang2018}
Xuan Wang, Huansheng Song, and Hua Cui.
\newblock Pedestrian abnormal event detection based on multi-feature fusion in
  traffic video.
\newblock {\em Optik}, 154:22--32, 2018.

\bibitem{CrowdBehaviorTripathi2019}
Gaurav Tripathi, Kuldeep Singh, and Dinesh~Kumar Vishwakarma.
\newblock Convolutional neural networks for crowd behaviour analysis: a survey.
\newblock {\em The Visual Computer}, 35(5):753--776, 2019.

\bibitem{SuspiciousPositionalRowe2005}
{\em {Detecting Suspicious Behavior From Only Positional Data With Distributed
  Sensor Networks}}, volume Volume 6: 5th International Conference on Multibody
  Systems, Nonlinear Dynamics, and Control, Parts A, B, and C of {\em
  International Design Engineering Technical Conferences and Computers and
  Information in Engineering Conference}, 09 2005.

\bibitem{BodyMovementBull1983}
P.~Bull.
\newblock {\em Body Movement and Interpersonal Communication}.
\newblock Wiley, 1983.

\bibitem{SituGoodall2019}
J.~R. {Goodall}, E.~D. {Ragan}, C.~A. {Steed}, J.~W. {Reed}, G.~D.
  {Richardson}, K.~M.~T. {Huffer}, R.~A. {Bridges}, and J.~A. {Laska}.
\newblock Situ: Identifying and explaining suspicious behavior in networks.
\newblock {\em IEEE Transactions on Visualization and Computer Graphics},
  25(1):204--214, 2019.

\bibitem{CaseBasedXia2015}
Li-min Xia, Bao-juan Yang, and Hong-bin Tu.
\newblock Recognition of suspicious behavior using case-based reasoning.
\newblock {\em Journal of Central South University}, 22(1):241--250, 2015.

\bibitem{ObjectDetectionLee2018}
J.~{Lee}, S.~{Lee}, and S.~{Yang}.
\newblock An ensemble method of cnn models for object detection.
\newblock In {\em 2018 International Conference on Information and
  Communication Technology Convergence (ICTC)}, pages 898--901, 2018.

\bibitem{IODEum2017}
S.~{Eum}, H.~{Lee}, H.~{Kwon}, and D.~{Doermann}.
\newblock Iod-cnn: Integrating object detection networks for event recognition.
\newblock In {\em 2017 IEEE International Conference on Image Processing
  (ICIP)}, pages 875--879, 2017.

\bibitem{GAPActionRecognitionWang2018}
Y.~{Wang}, W.~{Zhou}, Q.~{Zhang}, and H.~{Li}.
\newblock Convolutional neural networks with generalized attentional pooling
  for action recognition.
\newblock In {\em 2018 IEEE Visual Communications and Image Processing (VCIP)},
  pages 1--4, 2018.

\bibitem{HandwrittenTextNaur2018}
R.~R. {Nair}, N.~{Sankaran}, B.~U. {Kota}, S.~{Tulyakov}, S.~{Setlur}, and
  V.~{Govindaraju}.
\newblock Knowledge transfer using neural network based approach for
  handwritten text recognition.
\newblock In {\em 2018 13th IAPR International Workshop on Document Analysis
  Systems (DAS)}, pages 441--446, 2018.

\bibitem{First3DCNNJi2013}
S.~{Ji}, W.~{Xu}, M.~{Yang}, and K.~{Yu}.
\newblock 3d convolutional neural networks for human action recognition.
\newblock {\em IEEE Transactions on Pattern Analysis and Machine Intelligence},
  35(1):221--231, 2013.

\bibitem{DeepCascadeSabokrou2017}
M.~{Sabokrou}, M.~{Fayyaz}, M.~{Fathy}, and R.~{Klette}.
\newblock Deep-cascade: Cascading 3d deep neural networks for fast anomaly
  detection and localization in crowded scenes.
\newblock {\em IEEE Transactions on Image Processing}, 26(4):1992--2004, 2017.

\bibitem{ZhangGestures3DCNN2017}
L.~Zhang, G.~Zhu, P.~Shen, and J.~Song.
\newblock Learning spatiotemporal features using \hbox{3DCNN} and convolutional
  lstm for gesture recognition.
\newblock In {\em 2017 IEEE International Conference on Computer Vision
  Workshops (ICCVW)}, pages 3120--3128, 2017.

\bibitem{MRIAgeStimationUeda2019}
M.~{Ueda}, K.~{Ito}, K.~{Wu}, K.~{Sato}, Y.~{Taki}, H.~{Fukuda}, and T.~{Aoki}.
\newblock An age estimation method using 3d-cnn from brain mri images.
\newblock In {\em 2019 IEEE 16th International Symposium on Biomedical Imaging
  (ISBI 2019)}, pages 380--383, 2019.

\bibitem{MRIHyperactivityZou2017}
L.~{Zou}, J.~{Zheng}, C.~{Miao}, M.~J. {Mckeown}, and Z.~J. {Wang}.
\newblock 3d cnn based automatic diagnosis of attention deficit hyperactivity
  disorder using functional and structural mri.
\newblock {\em IEEE Access}, 5:23626--23636, 2017.

\bibitem{MRIAlzheimerKhagi2018}
B.~{Khagi}, C.~G. {Lee}, and G.~{Kwon}.
\newblock Alzheimer’s disease classification from brain mri based on transfer
  learning from cnn.
\newblock In {\em 2018 11th Biomedical Engineering International Conference
  (BMEiCON)}, pages 1--4, 2018.

\bibitem{Ref1ConvArchitecture2018}
Jihong Liu, Jing Zhang, Hui Zhang, Xi~Liang, and Li~Zhuo.
\newblock Extracting deep video feature for mobile video classification with
  elu-3dcnn.
\newblock In Benoit Huet, Liqiang Nie, and Richang Hong, editors, {\em Internet
  Multimedia Computing and Service}, pages 151--159, Singapore, 2018. Springer
  Singapore.

\bibitem{Ref3ConvArchitectures2019}
Noorkholis~Luthfil Hakim, Timothy~K. Shih, Sandeli~Priyanwada
  Kasthuri~Arachchi, Wisnu Aditya, Yi-Cheng Chen, and Chih-Yang Lin.
\newblock Dynamic hand gesture recognition using 3dcnn and lstm with fsm
  context-aware model.
\newblock {\em Sensors}, 19(24):5429, Dec 2019.

\end{thebibliography}

\end{document}